\documentclass[runningheads,a4paper]{llncs}

\usepackage{amsmath}
\usepackage{amssymb}
\usepackage{graphicx}
\usepackage{booktabs}
\usepackage{tabularx}
\usepackage{array}
\usepackage{ragged2e}

\usepackage{fontspec}
\usepackage{polyglossia}
\setmainlanguage{english}
\setotherlanguage{arabic}

\newfontfamily\arabicfont[
  Script=Arabic,
  Path=./,
  UprightFont=Amiri-Regular.ttf,
  BoldFont=Amiri-Bold.ttf,
  ItalicFont=Amiri-Italic.ttf,
  BoldItalicFont=Amiri-BoldItalic.ttf
]{Amiri-Regular.ttf}

\newcolumntype{L}{>{\RaggedRight\arraybackslash}X}
\newcolumntype{R}{>{\raggedleft\arraybackslash}X}
\begin{document}
\mainmatter

\title{Translation as a Decision Space: A Multi-Agent Perspective on Low-Resource Dialect Generation}

\author{
Hasan Alkhder\inst{1}\thanks{Corresponding author: hasan.alkhder2@ogr.sakarya.edu.tr} \and
Mohammad Abboush\inst{2} \and
Igor Tchappi\inst{3} \and
Ahmet Zengin\inst{1} \and
Amro Najjar\inst{4}
}

\institute{
Sakarya University, Turkey \and
TU Clausthal, Germany \and
University of Luxembourg, Luxembourg \and
Luxembourg Institute of Science and Technology, Luxembourg\\
\email{
hasan.alkhder2@ogr.sakarya.edu.tr\\
mohammad.abboush@tu-clausthal.de\\
igor.tchappi@uni.lu\\
azengin@sakarya.edu.tr\\
amro.najjar@list.lu
}
}

\date{} 
\maketitle

\begin{abstract}

Neural machine translation (NMT) systems typically produce a single output per input, obscuring the alternative decision trajectories implicitly available within multilingual decoding. This opacity becomes particularly problematic in low-resource dialect settings, where multiple linguistically valid realizations may differ in lexical authenticity, register, and structural stability.

We propose reframing translation as a structured decision space explored by autonomous translation agents. Instead of analyzing a single output, we model distinct translation pathways as agents operating over a shared multilingual backbone. Inter-agent divergence is treated not as error but as an interpretable behavioral signal.

We conduct an empirical study on Turkish--Syrian Arabic translation using three agents: (1) zero-shot direct translation, (2) dialect-stabilized translation via lightweight fine-tuning, and (3) pivot translation through English. Evaluation is performed on 5,000 dialogue sentences, while stabilization is trained on 5,000 additional Turkish--Syrian sentence pairs drawn from television dialogue and MADAR-Turk resources.

Rather than optimizing for conventional performance metrics, we quantify structured behavioral displacement using dialect marker frequency, lexical proximity to standardized Arabic, and structural variance. Lightweight stabilization nearly doubles dialect marker usage, increasing it from 0.2266 to 0.4988, while significantly reducing structural instability. Pivot mediation introduces normalization pressure and measurable compression effects, whereas zero-shot translation exhibits the highest decision variance.

We argue that translation divergence across agents reveals latent decision flexibility within multilingual models. By modeling translation as a multi-agent system over a shared decision space, we provide a principled interpretability framework for low-resource dialect generation.


\end{abstract}
\section{Introduction}

Neural machine translation (NMT) systems are commonly treated as deterministic functions that map a source sentence to a single target realization. In practice, however, multilingual generative models operate over high-dimensional probability distributions that encode multiple plausible translation trajectories. Despite this inherent flexibility, only one output surface form is typically exposed to the user. The alternative decisions embedded within the decoding space remain hidden. This creates a limitation for dialect translation analysis, because multiple valid realizations may exist while standard translation pipelines expose only a single output.

This opacity becomes especially consequential in low-resource dialect settings. Unlike standardized language translation, dialect translation involves competing realizations that differ not only lexically but socially and stylistically. Dialectal outputs may vary in register, authenticity, emotional intensity, and structural compression. In such contexts, divergence across outputs is not necessarily error; it may reflect meaningful trade-offs within the model’s internal decision landscape. Existing MT evaluation approaches capture translation adequacy reasonably well, but they are less suited for analyzing dialect authenticity, register choice, and structural variation across alternative translation realizations.

Syrian Arabic provides a particularly challenging case. It lacks standardized orthography, exhibits lexical divergence from Modern Standard Arabic (MSA), and is supported by limited parallel resources \cite{bouamor2018madar}. Multilingual models trained primarily on standardized data frequently display instability when generating dialectal text. This instability manifests as partial normalization toward MSA, repetition collapse, structural drift, or inconsistent colloquial weighting. Metrics such as BLEU \cite{papineni2002bleu} and COMET \cite{rei2020comet}, while effective for measuring translation adequacy, do not capture dialectal register, normalization pressure, or structural behavioral differences across translation pathways.

In this work, we propose reframing translation as a \emph{decision space} explored by translation agents. We adopt an operational definition of agents as constrained decoding pathways within a shared multilingual model, rather than fully autonomous entities in the classical multi-agent systems (MAS) sense. Each agent corresponds to a specific decision constraint applied to the same underlying model. Rather than treating translation as a single-output mapping, we model distinct translation pathways as independent agents operating over a shared multilingual backbone. Each agent embodies a specific decision constraint—zero-shot generation, dialect-reinforced adaptation, or pivot mediation—while preserving identical architecture and decoding parameters. We use the term agent because each pathway produces systematically different translation behavior under a defined constraint regime. The observable divergence between agents is interpreted not as noise, but as structured behavioral signal.

We conduct an empirical case study on Turkish–Syrian Arabic translation. Three agents are constructed:
(1) direct zero-shot Turkish $\rightarrow$ Syrian translation,
(2) dialect-stabilized translation via lightweight fine-tuning on 5,000 Turkish–Syrian sentence pairs, and
(3) pivot translation through English (Turkish $\rightarrow$ English $\rightarrow$ Syrian). Evaluation is performed on 5,000 aligned dialogue segments drawn from independent dialogue sources. This controlled setup allows us to isolate decision-space reweighting effects from architectural changes.

Instead of optimizing for conventional performance metrics, we quantify inter-agent divergence using dialect marker frequency, lexical proximity to standardized Arabic, and structural length variance. Together, these metrics capture complementary aspects of dialect activation, normalization pressure, and structural stability across translation pathways. Our findings show that lightweight dialect stabilization nearly doubles dialect marker usage while reducing structural instability, whereas pivot mediation introduces compression and normalization effects. Zero-shot translation exhibits the highest variance, reflecting under-specified dialect representation.

By modeling translation as a multi-agent system operating within a shared decision space, we introduce an interpretability-oriented framework for analyzing low-resource dialect generation. Rather than evaluating translation solely through final-output quality, our approach examines how controlled variation in translation pathways reveals structured behavioral differences within the same multilingual model. This perspective reframes divergence not as translation failure, but as an observable signal of how multilingual systems balance authenticity, normalization, and semantic trade-offs under different decision constraints. The central research question of this work is whether controlled translation pathway variation can reveal systematic and interpretable differences in low-resource dialect translation behavior.

Importantly, this study does not aim to improve machine translation accuracy. 
Instead, translation is used as a controlled generative environment in which 
agent-level behavioral divergence can be observed and analyzed.

The contributions of this work are threefold. 
First, we introduce a decision-space perspective for analyzing multilingual translation systems by modeling translation pathways as constrained agents. 
Second, we propose behavioral metrics to quantify inter-agent divergence in dialect activation, lexical normalization, and structural realization. 
Third, we provide an empirical analysis of Turkish–Syrian Arabic translation demonstrating how stabilization and pivot mediation reshape decoding behavior.

We emphasize that the notion of "agents" in this work is an abstract and analytical construct rather than a realization of classical multi-agent systems. 
The objective is not to model autonomous or interacting agents, but to provide a structured perspective for analyzing controlled decoding pathways within a shared multilingual model. 
In this sense, agents represent constrained operational views of the same system, enabling systematic exploration of the translation decision space.

\section{Related Work}

\paragraph{Transfer Learning for Low-Resource MT}

Transfer learning has become a widely adopted strategy for improving translation performance in data-scarce scenarios. Several studies have shown that knowledge acquired from high-resource language pairs can be transferred to low-resource tasks. For instance, Neubig et al. proposed similar-language regularization through joint training on related languages to reduce overfitting \cite{neubig2018rapid}. Gu et al. framed low-resource translation as a meta-learning problem that enables rapid adaptation from limited data \cite{gu2018meta}. Kim et al. investigated cross-lingual transfer using language-agnostic encoders and cross-lingual embeddings to alleviate vocabulary mismatch \cite{kim2019effective}, while Luo et al. introduced a hierarchical transfer framework that progressively trains on unrelated high-resource pairs, intermediate languages, and finally low-resource pairs \cite{luo2019hierarchical}. 

Although these techniques significantly improve translation quality, they generally produce deterministic single-output translations without revealing alternative translation hypotheses or underlying decision mechanisms.

\paragraph{Arabic Dialect Translation}

Arabic dialect translation introduces additional challenges due to morphological complexity, code-switching, and the limited availability of dialectal corpora. In this context, Sibaee et al. proposed SHAMI-MT, a bidirectional translation system for Syrian Arabic and Modern Standard Arabic based on AraT5v2 and trained on the Nabra dataset \cite{sibaee2025shami}. Baniata et al. developed a multitask learning architecture that shares a decoder across dialect translation tasks to exploit cross-task information \cite{baniata2018neural}. Salloum et al. demonstrated that unsupervised morphological segmentation can improve dialectal translation performance \cite{salloum2013dialectal}, while Shapiro et al. examined the influence of dialect identification on Arabic neural machine translation \cite{shapiro2019comparing}. 

Furthermore, the MADAR shared tasks have advanced dialect identification research, achieving macro-averaged F1-scores above 67\% across 25 dialects \cite{abbas2019st,talafha2019mawdoo3}. Alkheder et al. introduced the MADAR-Turk dataset, which contains manually translated Damascus–Turkish sentence pairs intended to facilitate benchmarking in dialectal translation tasks \cite{alkheder2023benchmarking}. 

Despite these developments, most approaches treat translation as a single-stage mapping process and provide limited interpretability regarding the influence of lexical and morphological variations on translation decisions.

\paragraph{Explainability in Neural Machine Translation}

Explainable AI techniques have also been explored to analyse the behaviour of neural translation models. Ferrando et al. proposed ALTI+, which traces token-level attribution across source and target representations \cite{ferrando2022towards}. Voita et al. adapted Layerwise Relevance Propagation for Transformer models to measure source and target contributions \cite{voita2021analyzing}. Moradi et al. showed that different attention distributions can produce identical predictions, highlighting the limitations of attention-based explanations \cite{madsen2022evaluating}. Additional studies have evaluated explanation fidelity and attention-head importance in translation models \cite{wang2021explanations}. 

However, these approaches mainly provide post-hoc interpretations of single outputs rather than exposing the broader decision processes underlying translation generation.

\paragraph{Multi-Agent Approaches and Pivot Translation}

Recent work has also explored multi-agent frameworks for machine translation. Wu et al. introduced TransAgents, a collaborative architecture inspired by human translation workflows \cite{wu2025perhaps}. Other studies proposed multi-agent learning strategies based on ensemble distillation, mutual learning, or collaborative debate mechanisms \cite{bi2019multi,liao2020multi,lai2023multidimensional}. 

While these methods improve translation quality through collaborative refinement, they typically treat agents as separate models rather than as entities exploring structured translation alternatives.

Pivot translation represents another approach for low-resource language pairs by exploiting high-resource intermediate languages. For example, Kim et al. proposed pivot-based transfer learning through step-wise pre-training \cite{kim2019pivot}. Additional work investigated multi-pivot ensembling and simultaneous multi-pivot translation strategies to improve robustness \cite{mohammadshahi2024investigating,dabre2019exploiting}. 

Despite their effectiveness, pivot-based approaches may propagate errors across translation stages and provide limited insight into the decision processes involved in pivot-mediated translation.

\paragraph{Summary and Research Gap}

Overall, the existing literature reveals an important research gap in neural machine translation for low-resource dialects. Most approaches focus on improving translation quality while treating translation as a deterministic mapping that generates a single output. Consequently, current methods offer limited insight into the internal decision processes of translation models or the alternative translation trajectories that may exist. This limitation is particularly critical for dialectal translation, where linguistic variability and limited data availability require more interpretable systems capable of explaining how lexical and structural choices are made during translation.

\section{Data Resources and Experimental Setup}

\subsection{Evaluation Dataset}

The evaluation corpus consists of 5K sentence-level dialogue segments aligned across multiple linguistic representations. The dataset is constructed as a fixed evaluation set rather than a benchmark-scale testbed. Since the goal of this study is controlled behavioral comparison between translation pathways, a stable set of 5K sentences provides sufficient coverage while enabling systematic analysis under identical conditions.

All translation agents operate on the same evaluation corpus so that observed behavioral differences can be attributed to translation pathway effects rather than dataset variation.

Each entry contains four aligned linguistic representations. The Turkish column (TR) contains the original source sentence. The Syrian Arabic column (SY) provides the reference dialectal translation derived from professionally dubbed dialogue. The English column (EN) contains an aligned English rendering used for pivot translation modeling. The Modern Standard Arabic column (MSA) provides a standardized Arabic form used for lexical proximity comparison.

Although the agents generate Turkish$\rightarrow$Syrian outputs exclusively, the presence of English and MSA columns enables structured analysis of pivot mediation and normalization effects.

The 5K evaluation sentences are drawn from two complementary sources. 
The first consists of 3K dialogue segments extracted from a professionally dubbed Turkish television series provided by NIS Studios, capturing naturally occurring conversational speech with emotional and informal discourse patterns. 

The second consists of 2K conversational sentence pairs from the MADAR-Turk corpus \cite{alkheder2023benchmarking}, providing structurally controlled conversational data.

Combining these sources balances natural dialogue richness with structurally controlled conversational data. Television dialogue contributes authentic colloquial phrasing and discourse markers, while the additional conversational data provides controlled coverage of common conversational patterns.

Importantly, the evaluation dataset is constructed independently and does not overlap with the stabilization (fine-tuning) data, ensuring that all reported results reflect generalization rather than memorization.

\paragraph{Example Entry.}

\begin{quote}
\textbf{TR:} Ben ümitli değilim açıkçası. \\
\textbf{EN:} I'm not hopeful, to be honest. \\
\textbf{MSA:} \textarabic{أنا لست متفائلًا بصراحة.} \\
\textbf{SY:} \textarabic{يعني أنا ماني كتير متفائلة بالموضوع.}
\end{quote}

This multi-column alignment enables relational comparison between dialectal output, standardized Arabic, and pivot-mediated translations without constraining agents to a single reference target.


\subsection{Dialect Stabilization Dataset}

To model dialect reinforcement as decision-space reweighting, we construct a separate fine-tuning dataset consisting of 5K Turkish–Syrian sentence pairs.

The stabilization corpus combines two sources as shown in Table~\ref{tab:stabilization}. The television dialogue portion was obtained from professionally dubbed Turkish television scripts provided by NIS Studios, a media production studio specialized in Turkish–Syrian Arabic dubbing.

\begin{table}[h]
\centering
\caption{Dialect Stabilization Dataset Composition}
\label{tab:stabilization}
\begin{tabular}{lc}
\toprule
Source & Number of Sentences \\
\midrule
Television Dialogue (NIS) & 3K \\
MADAR-Turk Conversational Pairs & 2K \\
\midrule
Total & 5K \\
\bottomrule
\end{tabular}
\end{table}

The MADAR-Turk corpus is used entirely for stabilization (fine-tuning), and no internal split is applied.

This composition enables effective dialect reinforcement while maintaining diversity in conversational structure and avoiding overfitting to a single narrative domain.


\subsection{Fine-Tuning Configuration}

Fine-tuning is intentionally lightweight. The objective is not to maximize end-task translation performance but to introduce controlled dialectal reinforcement within the existing multilingual representation space.

The model is fine-tuned on 5K Turkish–Syrian sentence pairs for a single epoch using a learning rate of $1 \times 10^{-5}$. No vocabulary extension or architectural modifications are applied, and all agents share identical decoding parameters.

This constrained training regime limits parameter updates substantially, ensuring that observed behavioral shifts primarily reflect lexical reweighting rather than large-scale structural retraining.


\subsection{Experimental Control and Separation}

Strict separation between evaluation and stabilization datasets is enforced. 
Although both datasets are constructed from similar sources (NIS dialogue and MADAR-Turk), 
they consist of distinct sentence instances with no overlap at the sentence level.

By controlling architecture, decoding parameters, and dataset separation, 
translation pathway remains the only varying factor across agents. 
This controlled setup enables systematic comparison of translation behavior 
within the proposed decision-space framework.


\section{Multi-Agent Translation Framework}

Neural machine translation systems are often described as mappings of the form $f(x)=y$, where a source sentence $x$ produces a target output $y$. This representation is a simplifying abstraction. In practice, multilingual generative models define a conditional probability distribution over possible outputs. For a given input $x$, decoding explores a high-dimensional distribution that contains multiple plausible realizations. The single output returned to the user therefore represents only one trajectory through a broader latent decision landscape.

We interpret this flexibility as a structured translation decision space. In this work, the decision space refers to the set of outputs generated when the same multilingual model is executed under different controlled pathway constraints. These outputs represent observable samples from the model’s conditional output distribution.

\begin{figure}[t]
\centering
\includegraphics[width=0.9\textwidth]{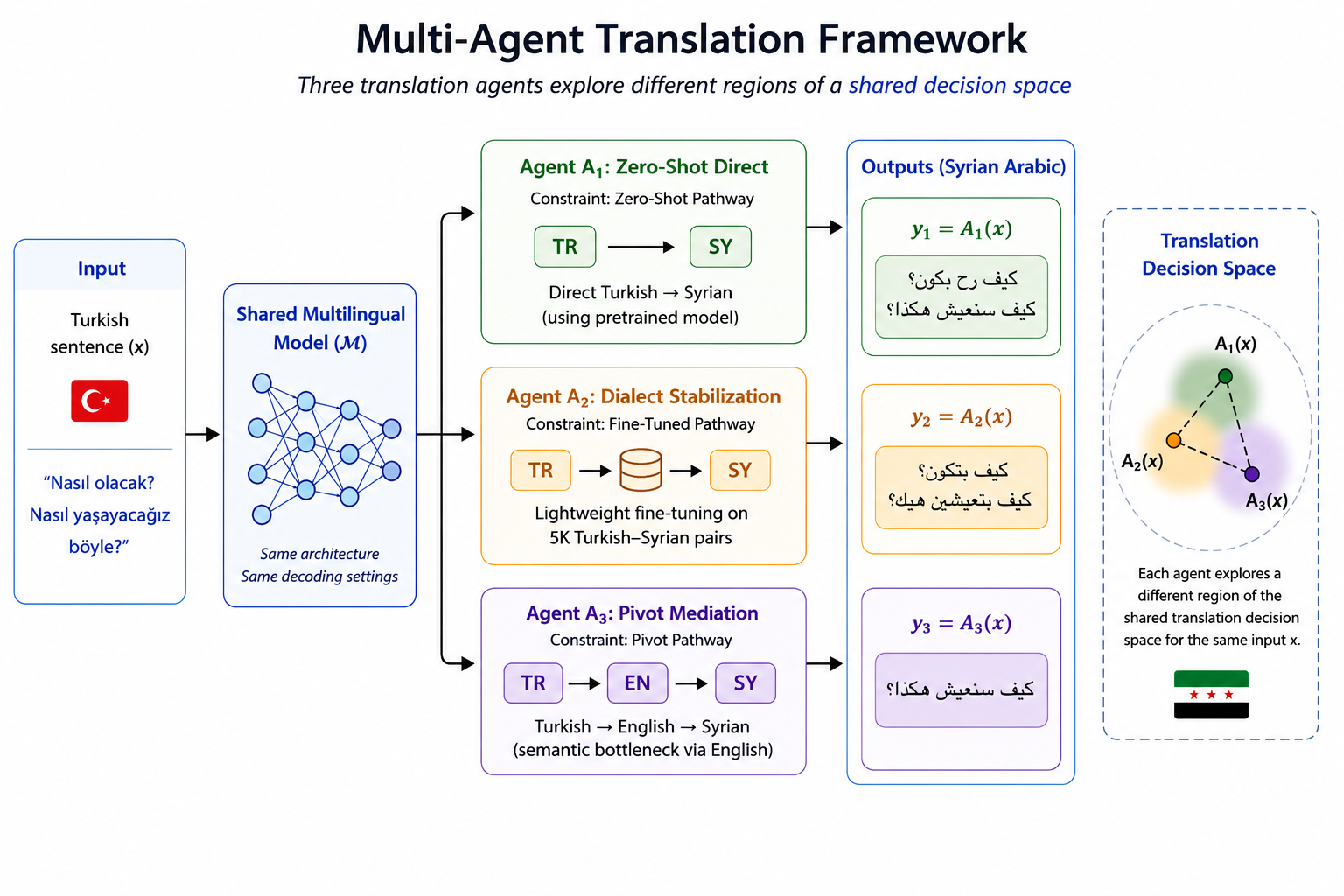}
\caption{
Overview of the proposed decision-space framework. A single multilingual model is executed under three constrained translation pathways (agents): zero-shot, stabilized, and pivot-based. The resulting outputs represent different regions of the shared translation decision space.
}
\label{fig:decision_space}
\end{figure}

As illustrated in Figure~\ref{fig:decision_space}, translation pathways define distinct regions within the shared decision space.

Formally, for a Turkish input sentence $x$, the multilingual backbone model defines a conditional distribution $P(y \mid x)$ over possible Syrian Arabic realizations. Rather than treating this distribution as an opaque sampling mechanism, we introduce controlled translation pathways that modify how decoding explores the distribution.

Each pathway is modeled as a translation agent representing a constrained decoding configuration of the same underlying model. We emphasize that agents in this framework are not fully autonomous entities in the classical MAS sense, but operational configurations defined by distinct decision constraints applied to a shared multilingual system.

Let $\mathcal{M}$ denote the fixed multilingual model. An agent $A_i$ maps an input sentence $x$ to an output $y_i$ under a specific operational constraint $C_i$:

\[
A_i(x) = \mathcal{M}(x; C_i).
\]

The constraint $C_i$ does not modify the architecture of $\mathcal{M}$ but specifies the translation pathway under which decoding is performed.

To analyze how translation behavior changes under different operational conditions, we examine three representative pathway configurations.

The first agent performs direct zero-shot Turkish→Syrian translation using the pretrained multilingual representation without dialect-specific adaptation.

The second agent introduces dialect stabilization through lightweight fine-tuning on 5K Turkish–Syrian sentence pairs. The fine-tuning is intentionally minimal so that the multilingual representation is not substantially retrained, but lexical activation patterns encouraging dialectal realization are slightly reinforced.

The third agent implements pivot mediation, decomposing translation into two sequential mappings: Turkish→English followed by English→Syrian. The intermediate English representation acts as a semantic bottleneck that can introduce additional normalization pressure on the final output.

All agents share identical decoding parameters and operate on the same evaluation dataset. The only variation lies in the pathway constraint $C_i$.

For each input $x$, we observe a subset of the translation decision space generated by the selected pathway configurations:

\[
\mathcal{D}(x) = \{ A_1(x), A_2(x), A_3(x) \}.
\]

This set represents observable realizations under the three controlled pathway constraints rather than the full space of outputs defined by $P(y|x)$.

The elements of $\mathcal{D}(x)$ are not interpreted hierarchically as correct versus incorrect translations. Instead, they represent neighboring realizations within the same semantic manifold. Divergence between agents therefore reflects behavioral displacement within the model’s latent representation space.

This perspective shifts interpretability from internal probing toward relational analysis of outputs. Instead of analyzing hidden representations directly, we observe how translations change when decoding conditions are modified in a controlled manner \cite{belinkov2019analysis}. In this framework, divergence across agents becomes an interpretable signal revealing how multilingual models balance dialectal authenticity, normalization pressure, and structural stability.

Importantly, the agents are not independent models but coordinated decoding configurations of the same multilingual system. The multi-agent perspective therefore introduces structured pathway variation rather than additional architectures.

Within low-resource dialect settings, this variation is particularly informative. Instability in zero-shot generation, normalization under pivot mediation, and reinforcement under dialect stabilization correspond to different observable regions of the translation decision space, allowing systematic analysis of dialect generation behavior without modifying the underlying model architecture.
\section{Behavioral Metrics for Structured Divergence}

Standard reference-based metrics remain useful for measuring translation adequacy. However, they are insufficient for capturing behavioral variation across translation pathways. In this work, conventional correctness-based evaluation is therefore complemented with behavior-oriented analysis that measures how translation outputs shift under different decoding conditions. Instead of evaluating a single output against a reference, we quantify structured divergence within the translation decision space introduced in Section~4.

These metrics are designed to provide interpretable signals at the behavioral level. 
Dialect marker frequency reflects lexical-level interpretability, lexical overlap captures normalization tendencies, and structural length variation reflects stability in generation behavior.

The selected metrics capture three complementary aspects of translation behavior: dialectal lexical activation, normalization toward standardized Arabic, and structural realization.

Let $x$ denote a Turkish source sentence and $y_i = A_i(x)$ the output of agent $A_i$.

\subsection{Dialect Marker Frequency}

Let $\mathcal{M}$ denote a predefined set of Syrian dialect markers curated from the dialogue corpus and validated by native speakers (e.g., \textarabic{شو، كتير، هيك، مو، لسا}).

Dialect activation for a sentence $y$ is defined as:

\[
D(y) = \frac{1}{|y|} \sum_{t \in y} \mathbf{1}(t \in \mathcal{M})
\]

Corpus-level activation for agent $A_i$ over dataset $\mathcal{X}$ is:

\[
\overline{D}_{A_i} =
\frac{1}{|\mathcal{X}|}
\sum_{x \in \mathcal{X}} D(A_i(x))
\]

Higher values indicate stronger dialect reinforcement in the generated translations.

\subsection{Lexical Proximity to Standardized Arabic}

Lexical proximity to Modern Standard Arabic (MSA) is used to estimate normalization pressure in the translation process. Since Syrian dialect diverges lexically from MSA, stronger overlap with standardized Arabic may indicate a tendency toward normalization rather than dialectal realization.

Given standardized reference $S(x)$, lexical overlap is defined as:

\[
O(y,S) =
\frac{| \text{tokens}(y) \cap \text{tokens}(S(x)) |}
{|\text{tokens}(S(x))|}
\]

Agent-level proximity is computed as:

\[
\overline{O}_{A_i} =
\frac{1}{|\mathcal{X}|}
\sum_{x \in \mathcal{X}} O(A_i(x), S(x))
\]

Lower overlap therefore indicates stronger dialectal divergence from standardized Arabic.

\subsection{Structural Length Ratio}

Structural realization is measured through relative sentence length compared with the standardized reference.

\[
L(y) = \frac{\ell(y)}{\ell(S(x))}
\]

Corpus mean for agent $A_i$:

\[
\overline{L}_{A_i} =
\frac{1}{|\mathcal{X}|}
\sum_{x \in \mathcal{X}} L(A_i(x))
\]

Structural variance is computed as:

\[
\sigma_{L,A_i} =
\sqrt{
\frac{1}{|\mathcal{X}|}
\sum_{x \in \mathcal{X}}
(L(A_i(x)) - \overline{L}_{A_i})^2
}
\]

Higher variance indicates instability in structural realization across generated translations.

\subsection{Inter-Agent Divergence}

To quantify behavioral displacement between agents, we measure divergence across the three behavioral dimensions introduced above: dialect activation, lexical normalization, and structural realization.

Relational displacement between agents $A_i$ and $A_j$ is defined as:

\[
\Delta_{ij} =
\left|
\overline{D}_{A_i} - \overline{D}_{A_j}
\right|
+
\left|
\overline{O}_{A_i} - \overline{O}_{A_j}
\right|
+
\left|
\overline{L}_{A_i} - \overline{L}_{A_j}
\right|
\]

This score measures behavioral divergence between translation agents under different pathway constraints.

The three components are combined with equal weight because they capture complementary aspects of translation behavior and operate on different behavioral dimensions rather than comparable numeric scales. Equal weighting therefore avoids privileging one behavioral dimension over the others.

\section{Quantitative Results}
\label{sec:quantitative_results}

Table~\ref{tab:quant_results} reports aggregate behavioral metrics across the three translation agents. 
Because all agents share the same multilingual backbone, observed differences reflect pathway-induced decision reweighting rather than architectural variation.

To assess the robustness of observed differences, we performed paired comparisons across agents using bootstrap resampling over the evaluation set. The observed differences in dialect marker frequency and length ratios remained stable across resampled subsets, indicating that the reported trends are not driven by dataset-specific effects.

\begin{table}[htbp]
\centering
\caption{Aggregate Behavioral Metrics Across Translation Agents}
\label{tab:quant_results}
\small
\renewcommand{\arraystretch}{1.1}
\begin{tabular}{lccc}
\toprule
Metric & Zero-shot & Stabilized & Pivot \\
\midrule
Avg. Dialect Markers & 0.2266 & 0.4988 & 0.2578 \\
Avg Overlap with MSA & 0.2133 & 0.1117 & 0.1114 \\
Std Overlap & 0.2530 & 0.1717 & 0.1788 \\
Avg Length Ratio & 1.5498 & 1.0332 & 0.9354 \\
Std Length Ratio & 2.8761 & 1.1834 & 1.0669 \\
\bottomrule
\end{tabular}
\end{table}

\begin{table}[htbp]
\centering
\caption{Representative Sentence-Level Divergence Across Translation Agents}
\label{tab:divergence_examples}
\small
\renewcommand{\arraystretch}{1.2}

\begin{tabular}{p{0.95\textwidth}}
\toprule

\textbf{Example 1} \\[2pt]
\textbf{TR:} Nasıl olacak? Nasıl yaşayacağız böyle? \\
\textbf{A1 Zero-shot:} \textarabic{كيف رح يكون؟ كيف سنعيش هكذا؟} \\
\textbf{A2 Stabilized:} \textarabic{كيف بتكون؟ كيف بتعيشين هيك؟} \\
\textbf{A3 Pivot:} \textarabic{كيف سنعيش هكذا؟} \\
\textbf{Observed Pattern:} Register Shift + Addressee Reframing \\

\midrule

\textbf{Example 2} \\[2pt]
\textbf{TR:} Hadi açın kitaplarınızı. Müzik Müzik Müzik... \\
\textbf{A1 Zero-shot:} \textarabic{افتحوا كتبكم. ميوزيكم ميوزيكم...} \\
\textbf{A2 Stabilized:} \textarabic{خلينا افتحوا كتبكم} \\
\textbf{A3 Pivot:} \textarabic{خلينا نفتح كتب الموسيقى...} \\
\textbf{Observed Pattern:} Token Noise + Contextual Normalization \\

\midrule

\textbf{Example 3} \\[2pt]
\textbf{TR:} Neredeydi şu bardaklar? \\
\textbf{A1 Zero-shot:} \textarabic{وين هيك الكوب؟} \\
\textbf{A2 Stabilized:} \textarabic{وين كانت الكوبات؟} \\
\textbf{A3 Pivot:} \textarabic{وين كانت النظارات؟} \\
\textbf{Observed Pattern:} Lexical Drift / Semantic Misalignment \\

\midrule

\textbf{Example 4} \\[2pt]
\textbf{TR:} O teyze benim yüzümden sana kızdı değil mi? \\
\textbf{A1 Zero-shot:} \textarabic{عمتك كتير غاضبة من أجلي، صح؟} \\
\textbf{A2 Stabilized:} \textarabic{عمة كتير غاضبة منك عشان أنا، صح؟} \\
\textbf{A3 Pivot:} \textarabic{هي عمّتك غاضبة من أجلي، صح؟} \\
\textbf{Observed Pattern:} Role Reassignment + Blame Shift \\

\midrule

\textbf{Example 5} \\[2pt]
\textbf{TR:} Eski düzen mi? \\
\textbf{A1 Zero-shot:} \textarabic{قديمي؟} \\
\textbf{A2 Stabilized:} \textarabic{النظام القديم؟} \\
\textbf{A3 Pivot:} \textarabic{النظام القديم؟} \\
\textbf{Observed Pattern:} Structural Completion + Nominal Recovery \\

\midrule

\textbf{Example 6} \\[2pt]
\textbf{TR:} Kabul ediyor musunuz iş teklifimi? \\
\textbf{A1 Zero-shot:} \textarabic{بتقبلوا عرضي للعمل؟} \\
\textbf{A2 Stabilized:} \textarabic{بتقبلوا عرض عملي؟} \\
\textbf{A3 Pivot:} \textarabic{بتقبل عرض وظيفتي؟} \\
\textbf{Observed Pattern:} Argument Structure Reweighting \\

\bottomrule
\end{tabular}

\end{table}

Dialect stabilization produces the strongest behavioral shift. 
Average dialect marker frequency increases from 0.2266 in zero-shot translation 
to 0.4988 after fine-tuning, indicating substantial lexical reweighting toward Syrian dialect forms. 
At the same time, structural variance decreases sharply 
($\sigma_L = 2.8761 \rightarrow 1.1834$), 
demonstrating improved stability in sentence realization. 
Lexical overlap with standardized Arabic decreases 
(0.2133 → 0.1117), confirming stronger dialect divergence from MSA.

The pivot pathway exhibits a different profile. 
Dialect activation remains moderate (0.2578), 
while the average length ratio drops below unity (0.9354), 
indicating structural compression relative to standardized references. 
Such compression often reflects normalization effects introduced by pivot mediation, where colloquial expansions or emphatic constructions in Syrian dialect are reduced when passing through an intermediate English representation. 
Variance is reduced compared to zero-shot but remains slightly higher than stabilized, 
consistent with semantic mediation through English acting as a regularizing constraint.

\begin{figure}[t]
\centering
\includegraphics[width=0.85\textwidth]{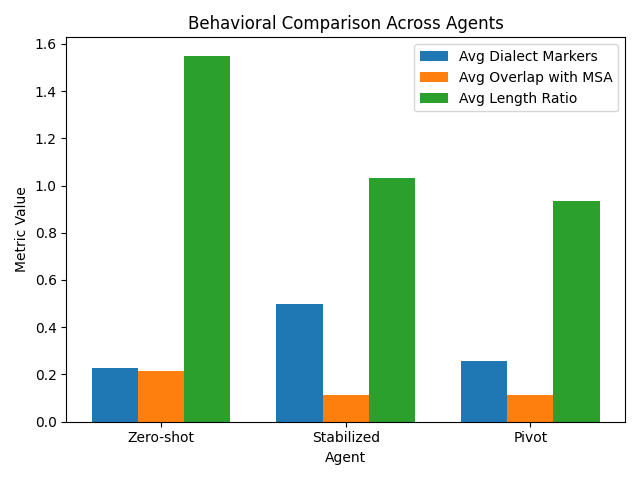}
\caption{
Aggregate behavioral comparison across translation agents. 
Stabilization amplifies dialect marker frequency and reduces structural expansion, 
while pivot mediation introduces compression and moderate normalization effects.
}
\label{fig:agent_comparison}
\end{figure}

As illustrated in Figure~\ref{fig:agent_comparison}, 
the three agents occupy distinct regions within the behavioral space. 
Zero-shot translation exhibits high variance and limited dialect activation. 
Stabilization shifts the system toward stronger dialect reinforcement and structural consistency. 
Pivot translation compresses output length and moderates lexical activation, 
revealing measurable behavioral differences consistent with decision-space reconfiguration across translation pathways.

Importantly, qualitative inspection of outputs indicates that increased dialectal activation does not appear to systematically degrade semantic adequacy. 
While minor lexical deviations occur, the core meaning is generally preserved across agents.

\section{Discussion: Decision-Space Reconfiguration Across Agents}

The empirical results reveal systematic behavioral displacement across translation agents. These shifts are not stochastic variations but structured reconfigurations within a shared multilingual decision space. Because all agents operate over the same backbone architecture, observed differences reflect pathway-induced reweighting rather than architectural divergence.

The zero-shot agent (A1) exhibits the highest structural variance ($\sigma_L = 2.8761$) and the lowest dialect marker activation (0.2266). This pattern suggests under-specified dialect representation: without explicit reinforcement, the model oscillates between standardized and colloquial realizations, resulting in unstable decoding trajectories. Zero-shot instability therefore reflects representational uncertainty rather than translation failure.

Dialect stabilization (A2) nearly doubles dialect marker frequency (0.4988) while significantly reducing structural variance ($\sigma_L = 1.1834$). Notably, this shift emerges under intentionally lightweight tuning—only a single epoch with a low learning rate—highlighting how even minimal dialect exposure can substantially reshape decoding behavior. Given this minimal adaptation regime, the observed change reflects lexical probability redistribution rather than structural retraining. Stabilization strengthens dialectal activation while preserving semantic integrity, demonstrating that low-resource dialect reinforcement primarily operates through probability reweighting within the existing representation space.

The pivot agent (A3) introduces an intermediate English representation and produces the lowest mean length ratio (0.9354) with moderate dialect activation (0.2578). This configuration imposes normalization pressure, frequently compressing emotionally loaded or colloquial constructions. Pivot mediation therefore acts as a regularizing constraint that reduces instability but attenuates dialect intensity.

Taken together, these results confirm that inter-agent divergence reflects controlled displacement in lexical weighting and structural realization. Translation pathways function as mechanisms for exploring neighboring regions of a common semantic manifold. Divergence is thus interpretable signal—revealing how multilingual systems negotiate authenticity, normalization, and mediation under different decision constraints.

\section{Sentence-Level Divergence Analysis}

While aggregate metrics capture behavioral displacement at the corpus level, 
sentence-level inspection reveals how these divergences manifest in concrete translation outputs. 
Examples were selected from sentences with the highest inter-agent divergence scores ($\Delta_{ij}$) as defined in Section~5.4. 
These cases therefore represent inputs where pathway-induced behavioral displacement across agents is most pronounced.

Table~\ref{tab:divergence_examples} reports the translations produced by the three agents for each selected sentence. 
Presenting the outputs side-by-side allows direct inspection of how different pathway constraints reshape lexical choice, syntactic realization, and semantic interpretation within the same underlying multilingual model.

Across examples, zero-shot outputs frequently exhibit lexical ambiguity, structural under-specification, or inconsistent register realization. Dialect stabilization increases colloquial activation while improving syntactic completeness. In contrast, pivot mediation often compresses emotionally marked constructions and occasionally introduces semantic reinterpretation through intermediate English normalization.

These qualitative observations are consistent with the quantitative patterns reported in Section~\ref{sec:quantitative_results}, confirming that inter-agent differences reflect structured behavioral variation rather than stochastic decoding noise.

\section{Implications for Explainable Multi-Agent Systems}

The proposed framework generalizes to generative decision systems where behavior can be interpreted through agent-level pathway variation.

Instead of probing internal activations or attention maps, interpretability emerges through relational comparison. Each agent operates under distinct constraints (direct, stabilized, pivot-mediated), and divergence between outputs becomes an explanatory mechanism.

In this context, explainability is not extracted post-hoc. It is constructed structurally through controlled variation of translation pathways. This approach is particularly valuable in low-resource settings, where ambiguity and variability are intrinsic rather than exceptional.

The multi-agent perspective also highlights that multilingual models contain latent decision flexibility. By selectively reinforcing or mediating translation pathways, we expose how lexical and structural weighting shifts within the same architecture.

\section{Limitations}

Several limitations should be acknowledged.

First, the study focuses exclusively on Turkish–Syrian Arabic translation. 
Although the decision-space framework is conceptually transferable, 
empirical validation across additional language pairs and dialect continua 
is required to confirm generalizability.

Second, dialect stabilization was intentionally lightweight 
(one epoch, low learning rate) to isolate lexical reweighting effects. 
More extensive adaptation may yield different structural dynamics, 
which were not explored in this study.

Third, the proposed behavioral metrics emphasize divergence rather than adequacy. 
While this aligns with our interpretability objective, 
future work may integrate performance-oriented measures 
to balance authenticity with translation quality.

Finally, lexical overlap calculations may be affected by orthographic variation 
in Arabic dialect writing. Although preprocessing was controlled, 
dialect spelling variability remains an inherent challenge in low-resource settings.

\section{Conclusion}

This study introduced a decision-space modeling framework for low-resource dialect translation, reframing translation pathways as constrained agents operating within a shared multilingual representation space. Rather than evaluating translation as a single deterministic output, we analyzed structured behavioral displacement across zero-shot, stabilized, and pivot-mediated configurations.

Empirical results over 5,000 evaluation sentences demonstrated that lightweight dialect stabilization substantially amplifies dialect marker activation while reducing structural instability. In contrast, pivot mediation introduces measurable normalization and compression effects. These systematic shifts confirm that translation pathways induce controlled reweighting within the translation decision space rather than stochastic variation.

By modeling translation as a multi-agent system, we show that inter-agent divergence can serve as an interpretability signal. Behavioral displacement across agents reveals how multilingual models balance dialectal authenticity, structural realization, and semantic mediation under different pathway constraints.

Beyond Turkish–Syrian translation, the proposed framework suggests a broader paradigm for analyzing generative systems through controlled pathway variation. Future work may extend this approach to additional dialect continua and generative tasks, further exploring pathway-based modeling as a mechanism for interpretable multilingual AI. More broadly, controlled pathway variation may serve as a practical strategy for explainable multi-agent analysis in generative language systems.

\section*{Acknowledgements}

This work was supported by the Luxembourg National Research Fund (FNR) through the project “The Epistemology of AI Systems” (C22/SC/17111440/EAI). The authors also thank NIS Studios for providing dialogue data used in the experiments of this study.

\bibliographystyle{splncs04}

\bibliography{custom}


\end{document}